\documentclass[11pt]{article}

\usepackage[T1]{fontenc}
\usepackage[utf8]{inputenc}
\usepackage[a4paper,margin=1in]{geometry}
\usepackage{lmodern}

\usepackage{amsmath}
\usepackage{amsfonts}
\usepackage{amssymb}
\usepackage{amsthm}
\usepackage{xparse}
\usepackage{stmaryrd}

\usepackage{graphicx}
\usepackage{xcolor}
\usepackage{booktabs}
\usepackage{longtable}
\usepackage{makecell}
\usepackage{array}
\usepackage{multirow}
\usepackage{float}

\usepackage[protrusion=true,expansion=false]{microtype}
\usepackage{nicefrac}
\usepackage{enumitem}

\usepackage[numbers,sort&compress]{natbib}
\usepackage{url}
\usepackage[hidelinks]{hyperref}

\usepackage{comment}

\usepackage{authblk}
\usepackage{orcidlink}

\theoremstyle{definition}

\theoremstyle{plain}

\theoremstyle{remark}

\NewDocumentCommand{\ECFP}{o}{%
\IfNoValueTF{#1}
{\mathrm{ECFP}}
{\mathrm{ECFP}_{#1}}%
}

\NewDocumentCommand{\ECFPrc}{o}{%
\IfNoValueTF{#1}
{\mathrm{ECFP}_{\mathrm{rc}}}
{\mathrm{ECFP}_{\mathrm{rc},#1}}%
}

\title{\textbf{An Agentic Retrobiosynthesis Framework with Learned Frontier Selection}}

\author[1]{Philippe Meyer\,\orcidlink{0000-0002-0618-2947}\thanks{Corresponding author: \texttt{philippe.meyer@inrae.fr}}}
\author[1]{Guillaume Gricourt\,\orcidlink{0000-0003-0143-5535}}
\author[1]{Thomas Duigou\,\orcidlink{0000-0002-2649-2950}}
\author[1]{Joan Hérisson\,\orcidlink{0000-0001-9741-0847}}
\author[1]{Jean-Loup Faulon\,\orcidlink{0000-0003-4274-2953}}

\affil[1]{{\normalsize Université Paris-Saclay, INRAE, AgroParisTech, Micalis Institute, Jouy-en-Josas, France}}

\date{\vspace{-1cm}}

\begin{document}
\maketitle

\begin{abstract}
Large language models are increasingly used as agents for multistep retrosynthesis, raising the question of how much their search policy contributes independently of the underlying reaction model. We investigate this question in a biological setting through rule-based retrobiosynthesis: a deterministic biochemical engine generates the same validated transitions for every method, searching for routes that terminate in metabolites available to an \emph{Escherichia coli} chassis, while the policy only selects which frontier molecule to expand next. Prompted and LoRA-tuned Qwen2.5-7B policies use a strict choice-only interface. The fine-tuned policy reaches $65\pm1$\% solve rate at 10 expansions on LASER versus 59\% for MCTS, and at 200 expansions reaches $78\pm1$\% versus 75\% on LASER, $88\pm3$\% versus 80\% on the RetroPath RL Golden benchmark, and $63\pm2$\% versus 45\% on the BioNavi-NP benchmark. Fine-tuning also consistently outperforms direct prompting. These results show that route-supervised frontier selection can improve budgeted search without altering biochemical generation, although performance remains dependent on frontier construction and reaction ranking.
\end{abstract}

\section{Introduction}
\label{sec:intro}

Retrobiosynthesis recursively decomposes target molecules through enzyme-catalyzed transformations. In metabolic engineering, RetroPath2.0, RetroPath RL, and BioNavi-NP search for routes terminating in metabolites available to a host chassis \citep{delepine2018retropath,koch2020rl,zheng2022bionavi}. Other frameworks are not host-specific: RetroBioCat designs biocatalytic cascades from enzymatic rules and literature precedent \citep{finnigan2021retrobiocat}, while EnzRetro couples enzymatic reaction generation and enzyme identification with MCTS \citep{cao2026enzretro}. Repeated biochemical expansion nevertheless remains a nontrivial multistep search problem \citep{gricourt2024artificial}.

Chemical retrosynthesis has recently become a test bed for LLM-based agents. Initial systems connected language models to chemistry tools \citep{zhang2024synask}, while more recent approaches have used LLMs to evaluate and steer candidate synthetic routes \citep{bran2026chemical,baker2025larc}. A further shift gives the model a more direct role in the planning process: RETRO-R1 learns an interactive policy from environment feedback \citep{liu2026retro}, Synthelite combines LLM-proposed synthesis strategies with MCTS \citep{xuan2025synthelite}, AOT* couples LLM-generated pathways with AND--OR search \citep{song2026aot}, and RETROAGENT gives the model structured access to the search state and expansion decisions \citep{zhu2026retroagent}.

\begin{figure}[t]
\centering
\includegraphics[width=0.9\linewidth]{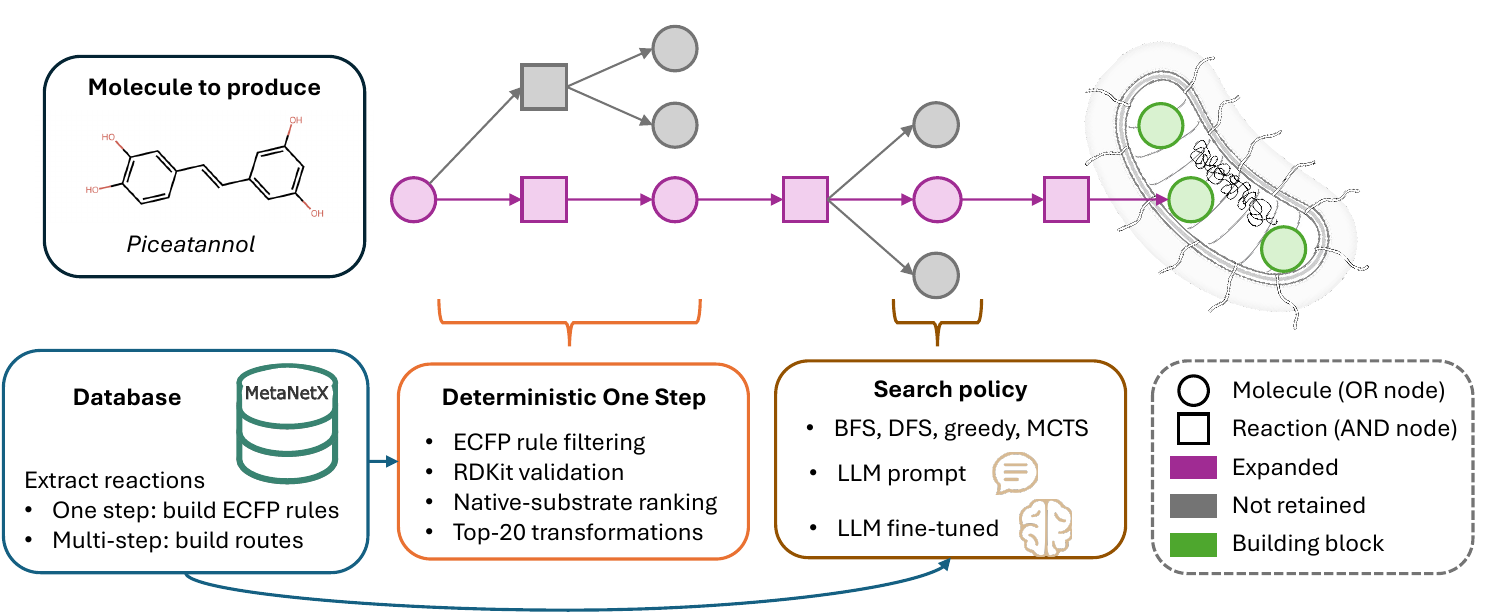}
\caption{\textbf{Overview of the agentic retrobiosynthesis framework.}
A target molecule is expanded through a deterministic MetaNetX-derived
one-step engine, while the search policy selects which frontier molecule to
expand next. The search proceeds toward the \emph{E.~coli} terminal metabolite
set.}
\label{fig:overview}
\end{figure}

Comparable agentic control remains largely unexplored in retrobiosynthesis.
This study isolates frontier selection as the only learned component while
holding biochemical reaction generation fixed, and compares prompted and
LoRA-fine-tuned Qwen2.5-7B-Instruct \citep{qwen2.5,qwen2, hu2022lora} against BFS, DFS,
greedy sink similarity, and MCTS \citep{kocsis2006bandit,segler2018planning}. Using a strict choice-only interface, the
experiments test whether supervision from mined biochemical routes improves
where a fixed expansion budget is spent, see Figure \ref{fig:overview}. Across three benchmarks, fine-tuning
consistently improves over direct prompting and achieves higher solve rates
than MCTS at the largest evaluated budget while leaving the underlying chemical
transitions unchanged.

\section{Framework}
\label{sec:framework}

\paragraph{Deterministic biochemical expansion.}
The one-step retrosynthesis model uses a reaction-center ECFP applicability
criterion \citep{rogers2010ecfp,meyer2026fingerprint} (Appendix~\ref{app:reaction_ecfp}) to screen 80,116 biochemical
reaction rules derived from MetaNetX
\citep{moretti2016metanetx,moretti2026metanetx}, discarding incompatible rules
through inexpensive vector operations before graph-level application. Rules
passing the filter are applied with RDKit \citep{landrum2013rdkit}, and
graph-valid outcomes are retained. The model uses ECFP radius $h=1$, with each
rule represented by a radius-2 ECFP-compatible reaction template. Reactions are
decomposed into mono-substrate transformations, and cofactors are removed from
precursor sets. The resulting candidate disconnections are ranked by Tanimoto similarity
between the query molecule and the native substrates from which their
biochemical rules were derived, providing a simple biochemical-precedent prior
similar to that used in RetroPath RL \citep{koch2020rl}. To control
combinatorial branching during multistep search, only the first 20 distinct
graph-validated disconnections are retained ($K_{\rm rxn}=20$).

\paragraph{Budgeted AND--OR search.}
The search graph contains molecule OR-nodes and reaction AND-nodes, with shared intermediates deduplicated across the graph. A molecule is solved when it belongs to the terminal set or when one of its child reactions is solved, whereas a reaction is solved only when all of its precursor molecules are solved. The terminal set contains 753 \emph{Escherichia coli} (\emph{E. coli}) metabolites derived from the RetroPath RL iML1515 sink \citep{koch2020rl,monk2017ml1515} (Appendix~\ref{app:ecoli_sink}), together with cofactors (Appendix~\ref{app:cofactors}). At step $t$, the search state is described by the graph $G_t$ and the unsolved frontier $F_t$. A policy selects a molecule $m_t \in F_t$, which is then expanded by the deterministic model. The search budget $N$ is defined as the number of molecule expansions.

\paragraph{Search policies.}
The non-LLM baselines are BFS, DFS, greedy sink similarity, and MCTS \citep{kocsis2006bandit,segler2018planning} with a
static sink-similarity evaluation. BFS and DFS prioritize minimum and maximum
graph depth, respectively. Greedy sink similarity selects the frontier molecule
with the highest Tanimoto similarity to any
metabolite in the terminal sink. MCTS uses the same sink-similarity measure as
a static value while accumulating search statistics through repeated
root-to-frontier descent. Complete policy definitions and parameters are given in
Appendix~\ref{app:search_policies}.

\paragraph{Agent observation.}
Unlike the non-LLM policies, which act on the complete frontier $F_t$, the LLM receives a bounded observation $V_t \subseteq F_t$ with $|V_t| \leq 20$ ($K_{\rm obs}=20$), to keep prompt size and inference cost manageable as the frontier grows. Similar bounded candidate views have been used in agentic chemical retrosynthesis to control inference cost and context size \citep{zhu2026retroagent}. The candidates included in $V_t$ are selected from the frontier using pre-expansion ranking signals derived from the current search state and reaction-rule applicability. We compare alternative constructions of this bounded observation using mined biochemical routes, as described below and in Appendix~\ref{app:ranking_diagnostics}. The displayed candidates are randomly shuffled before being passed to the LLM to mitigate systematic positional effects associated with option ordering \citep{pezeshkpour2024large}.

\paragraph{Mined routes and corpus construction.}
Following Retro* \citep{chen2020retro}, which learns search guidance from previously constructed retrosynthetic routes, we mine 12,251 biochemical routes from MetaNetX by tracing targets back to the \emph{E.~coli} building-block set and replay them through the search engine. These routes are used only for design and training; the three evaluation benchmarks are disjoint by construction. They serve two purposes. First, 2,000 routes yielding 4,262 decisions (3,821 with frontier size $>20$) are used to compare 26 frontier orderings for constructing $V_t$, with coverage@20 as the selection criterion (Appendix~\ref{app:ranking_diagnostics}). Second, the selected observation scheme is used to construct the route-replay corpus for imitation learning. Leakage filtering removes 1,198 pairs whose target is an evaluation target and 7,710 whose observation contains one, leaving 27,795 state--choice pairs, split by target into 25,150 training and 2,645 validation examples.

\paragraph{LLM frontier policies.}
We compare two Qwen2.5-7B-Instruct \citep{qwen2.5,qwen2} frontier policies using the same observation and action interface: the base instruction-tuned model used through direct prompting, and the same model fine-tuned on route-derived decisions with LoRA \citep{hu2022lora}. The fine-tuning corpus is obtained by replaying the mined biochemical routes: each training example pairs a bounded frontier observation with the frontier molecule lying on the reference route, which is used as the target expansion choice. At each decision, both policies receive the target SMILES, a compact summary of the current search graph, and up to 20 frontier candidates represented by their SMILES and graph depth; no reaction rule, EC annotation, or search history is provided.

\section{Experiments}
\label{sec:experiments}

Three retrobiosynthesis benchmarks are used. The RetroPath RL Golden benchmark contains 20 curated experimental pathways, corresponding to 70 reference one-step disconnections with known route lengths \citep{koch2020rl}. The LASER dataset contains 141 metabolic-engineering targets without reference routes \citep{winkler2015laser}. BioNavi-NP provides an independent natural-product benchmark \citep{zheng2022bionavi}. Due to computational cost, we uniformly sample 60 of its 388 usable test targets with a fixed seed (55 internal and 5 external cases), independently of model performance. All benchmarks are evaluated with the same \emph{E.~coli} terminal set and deterministic one-step model.

All policies use the same reaction rules, reaction-center filter, local
native-substrate ranking, top-20 disconnection cap, terminal set, depth limit,
graph update, and stopping condition. The LLM policies use the same
$K_{\rm obs}=20$ frontier observation on all three benchmarks. Performance is
measured as solve rate against the number of molecule expansions. LLM policies
are repeated across three different seeds, whereas deterministic
policies are evaluated once.

\section{Results}
\label{sec:results}

\paragraph{A four-signal portfolio maximizes coverage@20.}
Because the LLM can only act on displayed candidates, we select $V_t$ using
coverage@20 rather than MRR, prioritizing retention of at least one productive
choice after truncation. Across 3,821 states with more than 20 frontier
molecules, the selected four-signal portfolio reaches 84\% coverage@20, versus
80\% for depth stratification and 79\% for native-substrate similarity alone,
despite the latter having higher MRR (0.546 versus 0.505). In other words, the bounded observation retains at least one route-preserving frontier molecule in 84\% of states where truncation is required. The portfolio,
combining depth stratification, native-substrate similarity, reaction
precedent, and molecular size, is therefore used throughout the LLM experiments
(Appendix~\ref{app:ranking_diagnostics}).

\paragraph{Fine-tuning yields consistent gains across the large LASER benchmark.}
Table~\ref{tab:benchmark_search} reports solve rate as a function of the
expansion budget. On LASER, the fine-tuned Qwen2.5-7B policy reaches
$65\pm1$\% at $N=10$, compared with 59\% for MCTS, 55\% for BFS, and 53\%
for greedy sink similarity. The advantage persists as the budget grows:
$72\pm1$\% at $N=50$ and $78\pm1$\% at $N=200$, compared with 70\% and
75\% for MCTS. Direct prompting reaches $55\pm1$\% at $N=10$ and
$69\pm1$\% at $N=200$. Among successful runs, the fine-tuned policy reaches
a first solution after 12.6 expansions on average, compared with 17.8 for
prompting and 11.9 for MCTS. Its returned routes average 2.4 steps, compared
with 1.7 for prompting and 1.9 for MCTS.

\paragraph{Route supervision provides its strongest early-budget gain on Golden.}
On the 20 Golden targets, the fine-tuned policy reaches $68\pm8$\% at
$N=10$, compared with $40\pm5$\% for prompting, 45\% for MCTS, and 50\%
for BFS. The lead is maintained through the full budget curve, reaching
$77\pm6$\% at $N=50$ and $88\pm3$\% at $N=200$, compared with 75\% and
80\% for MCTS. At $N=200$, direct prompting reaches $70\pm5$\%. Among
successful runs, the fine-tuned policy reaches the first solution after 18.3
expansions on average, compared with 26.6 for prompting and 21.1 for MCTS.
Its solved routes average 3.3 steps, close to MCTS at 3.2 steps.

\paragraph{The learned policy transfers to the natural-product domain.}
On the BioNavi-NP target set, the fine-tuned policy reaches
$43\pm3$\%, $49\pm3$\%, $55\pm4$\%, $59\pm3$\%, and $63\pm2$\% for
$N=10,25,50,100,200$. At $N=200$, this is a 24-point gain over direct
prompting ($39\pm2$\%) and an 18-point gain over MCTS (45\%). The fine-tuned
policy reaches the first solution after 21.3 expansions on average, compared
with 38.6 for prompting and 25.0 for MCTS, while returning longer routes on
average (3.7 steps versus 2.6 and 2.7, respectively). The gain therefore generalizes across the complementary natural-product benchmark and persists through the largest evaluated budget.

\paragraph{Matched observations preserve the strongest classical baselines on Golden.}
As a control, we restrict the classical policies to the same bounded observation $V_t$ used by the LLM policies. On Golden at $N=200$, BFS and MCTS retain the same solve rates (75\% and 80\%, respectively), while DFS improves from 45\% to 50\% and greedy sink similarity from 45\% to 65\% (Appendix~\ref{app:matched_observation}). This suggests that the performance gain of the fine-tuned policy cannot be explained simply by restricting the search frontier to $V_t$.

\paragraph{Fine-tuning substantially reduces LLM time per successful route.}
At $N=200$, the fine-tuned policy requires 209, 168, and 417 seconds per obtained solution on LASER, Golden, and BioNavi-NP, respectively, compared with 854, 972, and 3069 seconds for direct prompting. Since both LLM policies use the same GPU/vLLM infrastructure, these values indicate a substantially lower campaign-level wall-clock cost for the fine-tuned policy. Comparisons with classical CPU-based policies should not be interpreted as intrinsic differences in algorithmic speed.

\begin{table}[t]
\centering
\caption{\textbf{Search performance on LASER, Golden, and BioNavi-NP.}
Solve rate (\%) is reported versus expansion budget. For LLM policies, solve
rates are the mean $\pm$ standard deviation over three random seeds; classical
policies are evaluated once. Exp.$\rightarrow$sol. is the mean number of
expansions required to reach the first solution over successful runs, and route
len. is the mean length of returned solved routes; both are measured at
$N=200$. Time/success is total wall-clock time, including failed runs, divided
by the number of successes at $N=200$; wall-clock values are hardware-dependent.}
\label{tab:benchmark_search}
\scriptsize
\resizebox{\linewidth}{!}{%
\begin{tabular}{lrrrrrrrr}
\toprule
Policy & $N{=}10$ & $N{=}25$ & $N{=}50$ & $N{=}100$ & $N{=}200$
& Exp.$\rightarrow$sol. & Route len. & Time/success (s)\\
\midrule
\multicolumn{9}{l}{\textit{LASER (141 targets)}}\\
BFS
& 55 & 69 & 69 & 72 & 74 & 12.4 & 1.9 & 362.0\\
DFS
& 55 & 56 & 56 & 56 & 57 & 4.0 & 1.7 & 482.1\\
Greedy sink similarity
& 53 & 55 & 56 & 57 & 57 & 3.8 & 1.7 & 973.8\\
MCTS
& 59 & 68 & 70 & 72 & 75 & 11.9 & 1.9 & 283.8\\
Qwen2.5-7B prompted
& 55$\pm$1 & 59$\pm$1 & 62$\pm$1 & 65$\pm$1 & 69$\pm$1
& 17.8 & 1.7 & 854.0\\
Qwen2.5-7B + LoRA
& \textbf{65$\pm$1} & \textbf{70$\pm$1} & \textbf{72$\pm$1}
& \textbf{74$\pm$2} & \textbf{78$\pm$1}
& 12.6 & 2.4 & 209.1\\
\midrule
\multicolumn{9}{l}{\textit{Golden (20 targets)}}\\
BFS
& 50 & 55 & 55 & 60 & 75 & 30.2 & 3.4 & 601.8\\
DFS
& 40 & 40 & 40 & 40 & 45 & 22.1 & 3.0 & 58.5\\
Greedy sink similarity
& 30 & 40 & 45 & 45 & 45 & 9.9 & 3.0 & 93.4\\
MCTS
& 45 & 65 & 75 & 75 & 80 & 21.1 & 3.2 & 23.1\\
Qwen2.5-7B prompted
& 40$\pm$5 & 48$\pm$3 & 57$\pm$8 & 65$\pm$0 & 70$\pm$5
& 26.6 & 2.7 & 972.0\\
Qwen2.5-7B + LoRA
& \textbf{68$\pm$8} & \textbf{75$\pm$9} & \textbf{77$\pm$6}
& \textbf{83$\pm$3} & \textbf{88$\pm$3}
& 18.3 & 3.3 & 168.2\\
\midrule
\multicolumn{9}{l}{\textit{BioNavi-NP (60 targets)}}\\
BFS
& 23 & 38 & 42 & 43 & 43 & 15.7 & 2.5 & 491.1\\
DFS
& 35 & 35 & 35 & 35 & 35 & 2.6 & 2.6 & 1046.1\\
Greedy sink similarity
& 17 & 17 & 18 & 18 & 18 & 6.6 & 3.1 & 7492.1\\
MCTS
& 25 & 32 & 38 & 43 & 45 & 25.0 & 2.7 & 1371.4\\
Qwen2.5-7B prompted
& 17$\pm$3 & 23$\pm$3 & 30$\pm$2 & 34$\pm$3 & 39$\pm$2
& 38.6 & 2.6 & 3069.0\\
Qwen2.5-7B + LoRA
& \textbf{43$\pm$3} & \textbf{49$\pm$3} & \textbf{55$\pm$4}
& \textbf{59$\pm$3} & \textbf{63$\pm$2}
& 21.3 & 3.7 & 416.9\\
\bottomrule
\end{tabular}%
}
\end{table}

\section{Discussion and Conclusion}
\label{sec:discussion}

Holding biochemical reaction generation fixed isolates frontier selection as the only learned component. Across all three benchmarks, route-supervised Qwen2.5 consistently improves over direct prompting and remains competitive with or better than MCTS throughout the evaluated search budgets. The advantage is already visible under tight budgets, indicating that the learned policy does not merely recover additional routes given more search, but allocates early expansions more effectively. At $N=200$, it reaches $78\pm1$\%, $88\pm3$\%, and $63\pm2$\% solve rates on LASER, Golden, and BioNavi-NP, respectively, versus 75\%, 80\%, and 45\% for MCTS. The particularly strong gain on BioNavi-NP further indicates that the learned search strategy transfers beyond the MetaNetX-derived training routes to a complementary natural-product benchmark.

Training itself is important. All policies operate over the same observation and action space and use the same biochemical reaction model, so performance differences reflect frontier-selection strategies rather than differences in reaction generation. Within the LLM policies, the fine-tuned and directly prompted variants additionally share the same base model, isolating the contribution of route supervision. Fine-tuning improves solve rate by 9--24 percentage points at $N=200$, with similarly substantial gains at smaller budgets, while also reducing the number of expansions required to reach a first solution.

This search-focused formulation is particularly relevant to synthetic biology. A solved retrobiosynthetic route is a design hypothesis connecting a target molecule to metabolites available in a host chassis, but its practical value additionally depends on enzyme availability and activity, thermodynamics, toxicity, flux, and compatibility with cellular physiology. Such biological information can also be incorporated during or after pathway generation: RetroPath RL combines chemical similarity with a biological score reflecting enzyme-sequence availability to guide and filter the search, BioNavi-NP supplements predicted routes with enzyme- and species-based annotations, RetroBioCat uses enzymatic precedent to rank biocatalytic pathways, and EnzRetro couples pathway prediction with enzyme identification
\citep{koch2020rl,zheng2022bionavi,finnigan2021retrobiocat,cao2026enzretro}. Efficient search can therefore reduce the combinatorial space while biological criteria help prioritize pathways for more detailed assessment and, ultimately, experimental testing.

The separation between deterministic biochemical generation and learned search also suggests a natural role for agentic methods. More broadly, LLM agents have been used to interleave reasoning with actions in external environments \citep{yao2022react} and to guide structured exploration through explicit planning and tree search \citep{hao-etal-2023-reasoning,pmlr-v235-zhou24r}. Feedback from previous trajectories can also improve subsequent decisions \citep{shinn2023reflexion}. In retrobiosynthesis, such agents could use the biological signals described above directly during frontier selection, making search increasingly chassis- and constraint-dependent. Recent agentic retrosynthesis systems similarly combine structured search with model-based reasoning and specialized tools \citep{bran2026chemical,zhu2026retroagent}, while retaining explicit control over the underlying transformations.

Several limitations remain. Matched-observation controls indicate that the bounded frontier view does not explain the learned policy's Golden advantage: strong classical policies are unaffected, while weaker heuristics can benefit from the restricted portfolio. Larger search budgets should nevertheless be investigated. Search success also depends on the reaction-rule system and chassis sink, while stereochemistry, thermodynamics, enzyme activity, yields, kinetics, and cellular physiology are not modelled. A solved route therefore establishes connectivity-level reachability to the selected \emph{E.~coli} chassis rather than an experimentally validated production pathway. Finally, while our experiments isolate the benefit of route supervision over prompting, comparisons with supervised non-LLM policies, as well as A*-like search with a learned cost-to-go heuristic as in Retro* and BioNavi-NP \citep{chen2020retro,zheng2022bionavi}, would help disentangle the contribution of the LLM architecture.

Overall, our results support a focused role for trained LLM agents in retrobiosynthesis: route supervision improves how a limited search budget is allocated without replacing the underlying biochemical model. More broadly, this suggests that the value of LLMs in pathway design may lie not only in their pretrained knowledge, but also in their ability to learn domain-specific planning strategies from successful scientific trajectories.

\phantomsection
\addcontentsline{toc}{section}{References}
\bibliographystyle{unsrtnat}
\bibliography{references}

\clearpage
\appendix

\section{Molecular representation and preprocessing}

All molecular structures are parsed with RDKit \citep{landrum2013rdkit} into two-dimensional molecular graphs and sanitized into canonical SMILES. Components containing wildcard atoms are discarded. Molecules are flattened to their two-dimensional graph representation, with stereochemical information removed. Unless stated otherwise, all molecular fingerprints, similarity computations, reaction-rule applications, and search operations use this representation.

\section{Reaction-center ECFP applicability filter}
\label{app:reaction_ecfp}

The reaction representation and applicability filter follow the fingerprint-space
construction of \citet{meyer2026fingerprint}; we summarize here the construction
used by the one-step model.

\paragraph{Biochemical reaction library.}
The one-step model is built from MetaNetX/MNXref version 4.5 compound and reaction tables \citep{moretti2016metanetx,moretti2026metanetx}, which reconcile metabolites and metabolic reactions across multiple biochemical databases. As MetaNetX reaction equations are represented without an assigned direction, each equation is instantiated in both orientations before atom mapping with RXNMapper\_v2 \citep{grandjean2026rxnmapperv2}. Unchanged components are removed, and reactions are decomposed into mono-substrate transformations by pairing each substrate with products sharing mapped atoms with it. Cofactors are removed from precursor sets.

\paragraph{Reaction templates.}
A reaction template is a local graph-rewrite rule extracted from an atom-mapped
reaction. It specifies the substrate pattern required for a transformation and
the corresponding product pattern around the reaction center. The reaction
center contains atoms whose local chemical signature changes between substrate
and product, or that have no product counterpart. The signature accounts for
formal charge, hydrogen count, valence, aromaticity, degree, ring membership,
and neighboring bond orders. For molecular ECFP radius $h$, templates retain
graph context up to radius $2h$ around the reaction center. The search engine
uses $h=1$, hence radius-2 reaction templates. Equivalent templates are merged,
yielding 80,116 templates. \citet{meyer2026fingerprint} report 70,037 templates for the same construction
after additionally discarding templates whose native substrate contains fewer
than five heavy atoms. This filter is not applied here because small metabolites
are legitimate precursors in retrobiosynthetic search.

\paragraph{Vector-space applicability filter.}
Applying every template to every frontier molecule would require repeated
subgraph matching over the full rule library. We therefore associate each
reaction $r$ with its reaction-center ECFP $\ECFPrc[h](r)$, which encodes the
local environments required around the reaction center. For a molecular system
$S$, a reaction is retained when
\begin{equation}
    \ECFPrc[h](r)+\ECFP_h(S)\geq0
\end{equation}
coordinate-wise. This is a necessary but not sufficient condition for
graph-level applicability: if an ECFP-compatible template applies to $S$, its
reaction-center ECFP necessarily passes this test, whereas the converse may
fail because ECFPs do not preserve complete molecular connectivity.

The criterion is therefore used only as a fast prefilter. All retained
templates are subsequently applied to the molecular graph with RDKit
\citep{landrum2013rdkit}, and only graph-valid products are returned by the
one-step engine. The filter reduces the number of expensive template
applications without changing the graph-level definition of a valid
retrosynthetic step.

\section{Cofactors and their treatment in the search}
\label{app:cofactors}

\paragraph{Cofactors.}
Cofactors are auxiliary species participating in biochemical transformations
that are not treated here as pathway-specific precursors to be synthesized.
The experiments use the biochemical cofactor set distributed with RetroRules
2026 \citep{duigou2026retrorules}, comprising 68 entries with InChI \citep{heller2015inchi}, InChIKey
connectivity prefixes, and compound names (Table~\ref{tab:cofactor_examples}).
The set includes classical coenzymes and redox carriers, ubiquitous small
molecules, and inorganic ions, such as ATP/ADP, NAD$^+$/NADH,
NADP$^+$/NADPH, CoA, water, phosphate, and common metal ions.

\paragraph{Treatment during search.}
When a validated reaction generates a pathway precursor together with a
cofactor, the cofactor is not added as an unresolved molecule to the frontier.
This prevents common species such as ATP, NADH, water, or phosphate from
creating additional branches in the AND--OR graph. A solved route therefore
assumes availability of the listed cofactors.

\begin{table}[t]
\centering
\caption{Representative entries from the biochemical cofactor list used in the
search. The complete list contains 68 records.}
\label{tab:cofactor_examples}
\small
\begin{tabular}{ll}
\toprule
Type & Representative entries \\
\midrule
Redox carriers & NAD$^+$/NADH; NADP$^+$/NADPH; FAD/FADH$_2$; FMN/FMNH$_2$ \\
Energy / phosphate transfer & ATP; ADP; AMP; GTP; UTP; phosphate; diphosphate \\
Group transfer & CoA; S-adenosyl-L-methionine; UDP; GDP; CDP \\
Small ubiquitous species & H$^+$; H$_2$O; O$_2$; CO$_2$; NH$_3$/NH$_4^+$ \\
Ions & Mg$^{2+}$; Fe$^{2+/3+}$; Zn$^{2+}$; Mn$^{2+/3+}$; Na$^+$; K$^+$ \\
\bottomrule
\end{tabular}
\end{table}

\section{\emph{Escherichia~coli} metabolites used as terminal building blocks}
\label{app:ecoli_sink}

\paragraph{Sink definition.}
A retrosynthetic branch is complete when all unresolved precursors belong to a
set of compounds assumed to be available in the host organism. This set is
referred to as the sink or terminal building-block set. Sink membership
therefore represents an assumption about chassis availability rather than
metabolite abundance under a particular growth condition.

\paragraph{Origin of the \emph{E.~coli} sink.}
The metabolite set is inherited from RetroPath RL \citep{koch2020rl}, where
organism-specific sinks are constructed from genome-scale metabolic models.
Cytosolic metabolites are retained and dead-end compounds that cannot be
produced in the steady-state model are removed using flux-variability analysis.
Chemical structures are then recovered from database cross-references and
standardized. The experiments use the sink derived from the \emph{E.~coli}
iML1515 genome-scale model \citep{monk2017ml1515}.

\paragraph{Processing.}
The initial collection contains 963 metabolite records, of which 809 produce a
valid InChIKey connectivity block. Deduplication at this level gives
753 unique terminal metabolites. Search molecules are matched using
the first 14-character InChIKey block rather than exact SMILES. This criterion
is insensitive to stereochemistry, isotopes, and charge or protonation variants
that are not explicitly modeled by the current search representation.

\section{Frontier ranking diagnostics}
\label{app:ranking_diagnostics}

The LLM acts on a bounded observation $V_t\subseteq F_t$ with
$K_{\rm obs}=20$. We therefore evaluate alternative frontier orderings by how
well they retain on-route molecules within this fixed 20-candidate
observation; the diagnostic does not optimize the observation size.

\paragraph{Metrics.}
For each replay state $j$, let $r_j\geq1$ denote the best rank among its
on-route frontier molecules. Let $J$ be the number of replay states and
$\mathcal{J}_k=\{j:|F_j|>k\}$ the subset for which a top-$k$ observation
actually truncates the frontier. We define
\begin{equation}
\operatorname{MRR}=\frac{1}{J}\sum_{j=1}^{J}\frac{1}{r_j},
\qquad
\operatorname{cov@}k=
\frac{100}{|\mathcal{J}_k|}
\sum_{j\in\mathcal{J}_k}\mathbf{1}[r_j\leq k].
\label{eq:frontier_metrics}
\end{equation}
Coverage@$k$ is thus the percentage of truncated states in which at least one
productive choice remains visible. MRR instead emphasizes how highly on-route
choices are ranked across all replay states. Since the LLM can only select
among the displayed candidates, coverage@20 is the primary selection
criterion. We also report the median rank over all replay states.

\paragraph{Ordering signals.}
All tested signals are available before graph expansion. Shallow- and
deep-depth orderings sort directly by graph depth. Depth stratification instead
groups frontier molecules by depth and traverses these groups round-robin,
preserving discovery order within each depth. A related ordering uses the same
depth stratification but ranks molecules within each depth by native-substrate
similarity.

Native-substrate similarity is the maximum Tanimoto similarity to the native
substrates associated with applicable MetaNetX rules. We also consider the
mean of the three highest such similarities and similarity breadth, defined as
the number of rules whose similarity is at least 90\% of the maximum. Reaction
precedent is the largest number of source MetaNetX reactions supporting any
applicable rule. Other signals include sink closeness, similarity to the
original target, the number of applicable rules, and molecular size measured
by heavy-atom count. The creation-order baseline simply preserves the order in
which frontier nodes were added to the graph.

We compare 26 individual and composite orderings. Some composites combine
scores algebraically, for example by penalizing native-substrate similarity
with depth or target similarity. A second family combines complete ranked
lists as portfolios. Given several orderings, the portfolio traverses them
round-robin: it takes the first candidate from each list, then the second from
each, and so forth. A molecule appearing in several lists is retained only at
its first occurrence, and later-ranked candidates fill the resulting free
positions. Thus, for the selected four-member portfolio, a 20-candidate view
would contain five candidates from each ordering in the absence of overlap,
but no fixed 5/5/5/5 quota is imposed when the rankings select the same
molecules. This construction allows complementary signals to contribute their
highest-ranked candidates without collapsing them into a single global score.

The selected portfolio interleaves depth stratification, native-substrate
similarity, reaction precedent, and increasing molecular size. These signals
respectively favor diversity across search depths, similarity to known
biochemical substrates, transformations with stronger MetaNetX precedent, and
smaller intermediates.

\paragraph{Frontier ordering.}
The diagnostic replays 2,000 attested routes and records 4,262 frontier
decisions. The replay always follows an on-route molecule independently of the
ordering being evaluated, so all orderings are compared on the same search
states. At $K_{\rm obs}=20$, 3,821 of these states satisfy $|F_t|>20$ and
therefore contribute to coverage@20.

Table~\ref{tab:frontier_orderings_full} reports the complete diagnostic. The
selected portfolio reaches 84\% coverage@20. In contrast, the highest-MRR
configurations reach only 77--78\% coverage@20, showing that higher MRR does
not necessarily retain more productive choices under top-20 truncation.

Depth stratification alone reaches 80\% coverage@20. The creation-order
baseline and shallow-depth ordering yield identical results (MRR 0.524 and
77\% coverage@20), indicating that molecule creation already carries a
shallow-depth bias. Sink closeness reaches only 43\% coverage@20, further
suggesting that proximity to the terminal metabolite set alone is a weak
criterion for constructing the LLM frontier observation.

\begin{table}[t]
\centering
\caption{Frontier-ordering diagnostic at $K_{\rm obs}=20$. MRR and median rank
are computed over all 4,262 replay decisions; coverage@20 is computed over the
3,821 states with $|F_t|>20$, for which the observation truncates the frontier.}
\label{tab:frontier_orderings_full}
\scriptsize
\setlength{\tabcolsep}{4pt}
\begin{tabular}{lrrr}
\toprule
Ordering & MRR & Median & Coverage@20 \\
\midrule
Depth / similarity / precedent / size portfolio
    & 0.505 & 3 & \textbf{84\%} \\
Depth / similarity portfolio
    & 0.533 & 2 & 83\% \\
Depth-weighted similarity / precedent portfolio
    & 0.518 & 2 & 82\% \\
Depth / similarity / precedent portfolio
    & 0.509 & 3 & 82\% \\
Native similarity within depth strata
    & 0.538 & 2 & 81\% \\
Depth stratification
    & 0.521 & 2 & 80\% \\
Native-substrate similarity
    & 0.546 & 2 & 79\% \\
Similarity / $(1+$depth$)$
    & 0.548 & 2 & 78\% \\
Similarity $-\,0.05\times$depth
    & 0.546 & 2 & 78\% \\
Similarity $\times$ mean-3 similarity
    & \textbf{0.553} & 2 & 77\% \\
Shallow depth
    & 0.524 & 2 & 77\% \\
Creation order
    & 0.524 & 2 & 77\% \\
Mean-3 native similarity
    & 0.550 & 2 & 77\% \\
Similarity $\times$ deep-depth factor
    & 0.438 & 3 & 71\% \\
Small molecular size
    & 0.328 & 8 & 66\% \\
Similarity $\times$ sink closeness
    & 0.320 & 7 & 65\% \\
Deep depth
    & 0.422 & 4 & 61\% \\
Similarity breadth
    & 0.294 & 12 & 54\% \\
Target similarity
    & 0.217 & 17 & 49\% \\
Sink closeness
    & 0.204 & 21 & 43\% \\
Similarity $-$ target similarity
    & 0.221 & 22 & 43\% \\
Reaction precedent
    & 0.205 & 23 & 42\% \\
Few applicable rules
    & 0.146 & 26 & 37\% \\
Similarity $\times(1-$target similarity$)$
    & 0.191 & 36 & 34\% \\
Many applicable rules
    & 0.130 & 36 & 26\% \\
Target dissimilarity
    & 0.131 & 53 & 23\% \\
\bottomrule
\end{tabular}
\end{table}

\section{Search policies and agent interface}
\label{app:search_policies}

\paragraph{Shared interface.}
All policies operate on the same evolving AND--OR graph. At step $t$, the
engine constructs the frontier $F_t$ of unsolved, unexpanded, non-terminal
molecule nodes. The policy returns one molecule $m_t\in F_t$ for expansion.
Reaction generation, applicability filtering, graph rewriting, chassis tests,
and solved-status propagation are unchanged across policies. The common search
budget is therefore the number of expanded molecules.

\paragraph{Classical traversal and sink-similarity policies.}
BFS selects a frontier molecule at minimum depth, whereas DFS selects one at
maximum depth. Greedy sink similarity uses
\begin{equation}
s_{\mathrm{sink}}(M)=\max_{B\in\mathcal{B}}
\operatorname{Tan}\!\left(\phi(M),\phi(B)\right),
\end{equation}
where $\mathcal{B}$ is the terminal chassis set, and expands the molecule with
maximum $s_{\mathrm{sink}}$. This baseline uses only structural proximity to an
available chassis metabolite and does not account for the remaining explored
route.

\paragraph{MCTS search.}

MCTS performs repeated root-to-frontier descent from the target through the explored
AND--OR graph \citep{kocsis2006bandit,segler2018planning}. For a child $i$ of a parent visited $N$ times, selection uses
\begin{equation}
\bar Q_i + c\sqrt{\frac{\log(N+1)}{1+n_i}},
\end{equation}
with exploration constant $c=1.0$, where $n_i$ is the child visit count and
$\bar Q_i$ its mean accumulated value. Each decision starts from the root and
alternates molecule OR-nodes and reaction AND-nodes until an eligible frontier
molecule is reached, with a maximum descent length of 64. Root-based descent is necessary because molecule nodes are expanded at most
once and then leave the frontier; applying the MCTS selection statistic directly to frontier nodes would
therefore provide no meaningful visit statistics. Repeated root-to-frontier
descent allows statistics to accumulate on shared search paths.

No rollout or learned value network is used. Instead, unvisited molecule nodes
are initialized by their sink closeness, computed as the maximum counted
Tanimoto similarity between their radius-1 ECFP and the metabolites in the
terminal sink. At an OR-node, an unvisited reaction is initialized by the
minimum sink closeness of its precursors, reflecting the AND requirement that
all precursors must be solved. At an AND-node, solved and terminal precursors
are skipped.

After the selected molecule is expanded by the common one-step engine, its
value is evaluated from the newly exposed local graph and backpropagated only
along the descent path. If the descent reaches a dead end without selecting an
eligible frontier molecule, that path receives zero value and the policy falls
back to the frontier molecule with maximum sink closeness.

\paragraph{Prompted and fine-tuned LLM policies.}
The LLM policies use the same action definition and do not generate reactions,
select templates, or modify the graph. Because the complete frontier may be
large, the model receives the bounded portfolio $V_t\subseteq F_t$, constructed
from depth stratification, native-substrate similarity, biochemical precedent,
and molecular size. It then returns the index of one displayed molecule. Direct prompting uses
Qwen2.5-7B-Instruct \citep{qwen2.5,qwen2} without additional training. The fine-tuned policy uses the
same base model with LoRA adapters trained on replayed state--choice pairs \citep{hu2022lora}.
Consequently, the comparison measures whether route-derived supervision
improves frontier selection while leaving chemical generation unchanged.

\begin{table}[t]
\centering
\caption{Search policies and the frontier observation available to each method.
Classical policies act on the complete frontier $F_t$, whereas LLM policies
select from the bounded observation $V_t$.}
\label{tab:policy_details}
\small
\begin{tabular}{lll}
\toprule
Policy & Selection signal & Observation \\
\midrule
BFS & minimum graph depth & $F_t$ \\
DFS & maximum graph depth & $F_t$ \\
Greedy sink similarity & maximum sink similarity & $F_t$ \\
MCTS & MCTS descent + sink value & $F_t$ \\
Qwen2.5-7B prompted & language-model ranking & $V_t$ \\
Qwen2.5-7B + LoRA & supervised language-model ranking & $V_t$ \\
\bottomrule
\end{tabular}
\end{table}

\section{LLM policy and supervised fine-tuning}
\label{app:llm_training}

\paragraph{Decision interface.}
The LLM is used exclusively as a search policy. At step $t$, the bounded
frontier $V_t\subseteq F_t$ is rendered as a list of candidate molecules and
the model returns only the index of the molecule selected for expansion, as the
strict JSON object \texttt{\{"choice": N\}}. No justification or explanatory
field is requested. Reaction generation, template application, graph updates,
and terminal tests are performed by the deterministic engine. The model
therefore cannot generate a reaction or directly modify the search graph.
Prompted and fine-tuned policies use the same observation and action format.

\paragraph{Training examples.}
Training examples are obtained by replaying 12,251 biochemical routes mined
from MetaNetX with the same search engine used for evaluation. Each example
contains the rendered search state and a frontier decision that preserves an
attested route. When several displayed molecules are compatible with the route,
the branch with the largest remaining route cost is selected by a fixed
convention. The target is therefore an imitation label rather than a claim of
unique optimality.

The frontier representation materially changes how much supervision can be
recovered from the same routes. The production four-signal portfolio yields
36,703 replay decisions, or 3.00 per mined route, and loses 26\% of otherwise
available decisions to truncation. Depth stratification alone yields 31,738
decisions (2.59 per route) and loses 37\%. Complete route replays are retained
for 56\% and 47\% of trajectories, respectively. The portfolio therefore
provides 16\% more replay decisions before benchmark filtering.

\begin{table}[h]
\centering
\caption{Effect of the frontier view on supervision recovered from the same
12,251 mined routes, before benchmark-target filtering.}
\label{tab:corpus_view}
\small
\begin{tabular}{lrrrr}
\toprule
Frontier view & Pairs & Pairs/route & Truncated away & Complete routes \\
\midrule
Portfolio & 36,703 & 3.00 & 26\% & 56\% \\
Depth stratification & 31,738 & 2.59 & 37\% & 47\% \\
\bottomrule
\end{tabular}
\end{table}

Benchmark molecules are removed by molecular skeleton before training.
Examples are also discarded when an evaluation target occurs in the displayed
frontier, preventing indirect overlap through routes mined for other targets.
After filtering, 27,795 state--choice pairs remain. Splitting by target gives
25,150 training examples and 2,645 validation examples over 709 held-out
validation targets. Decisions from the same route target therefore do not occur
in both sets.

\paragraph{Fine-tuning.}
Qwen2.5-7B-Instruct is fine-tuned using LoRA adapters of rank 16 on all linear
layers ($\alpha=32$, dropout $0.05$). Training is performed for one epoch with
a learning rate of $10^{-4}$, bfloat16 precision, gradient checkpointing, a
per-device batch size of 2, eight gradient-accumulation steps, and a maximum
sequence length of 3072 tokens. The resulting effective batch size is 16. The
loss is applied only to the assistant completion, while the frontier prompt is
used as context. Only the strict \texttt{\{"choice": N\}} completion is
supervised; no justification or explanatory reasoning is used as a training
target.

On the held-out validation split, which is not used for optimization or model
selection, the final model obtains a cross-entropy of 0.158 and a mean
token-level accuracy of 0.946, close to the corresponding training values.
These metrics assess imitation of the supervised frontier-choice format rather
than end-to-end search performance.

\begin{table}[h]
\centering
\caption{Fine-tuning configuration and training/validation metrics for the
learned frontier policy.}
\label{tab:llm_training}
\small
\begin{tabular}{lcc}
\toprule
\multicolumn{3}{l}{\textit{Fine-tuning configuration}} \\
\midrule
Component & \multicolumn{2}{l}{Setting} \\
\midrule
Base model & \multicolumn{2}{l}{Qwen2.5-7B-Instruct} \\
Training objective & \multicolumn{2}{l}{supervised frontier-choice imitation} \\
Filtered corpus & \multicolumn{2}{l}{27,795 state--choice pairs} \\
Train / validation & \multicolumn{2}{l}{25,150 / 2,645} \\
Held-out validation targets & \multicolumn{2}{l}{709} \\
LoRA rank / alpha / dropout & \multicolumn{2}{l}{16 / 32 / 0.05} \\
Target modules & \multicolumn{2}{l}{all linear layers} \\
Epochs & \multicolumn{2}{l}{1} \\
Learning rate & \multicolumn{2}{l}{$10^{-4}$} \\
Effective batch size & \multicolumn{2}{l}{16} \\
Maximum sequence length & \multicolumn{2}{l}{3072 tokens} \\
\midrule
\multicolumn{3}{l}{\textit{Training and validation metrics}} \\
\midrule
Metric & Train & Validation \\
\midrule
Cross-entropy, final logged step & 0.139 & 0.158 \\
Cross-entropy, epoch mean & 0.200 & -- \\
Mean token accuracy & 0.950 & 0.946 \\
Entropy & 0.140 & 0.139 \\
Runtime & 6 h 28 min & 12 min \\
\bottomrule
\end{tabular}
\end{table}
\paragraph{Inference and prompting baseline.}
During inference, a new frontier observation is generated for every
non-trivial decision and the model returns only \texttt{\{"choice": N\}}.
No model call is required when only one frontier molecule is available. The
prompting baseline uses the same Qwen2.5-7B-Instruct model and search interface
without LoRA fine-tuning. The system message and user-turn format are identical
for the prompted and fine-tuned policies; their exact text is given in
Appendix~\ref{app:llm_prompts}. The comparison therefore isolates the effect of
route-derived supervision on frontier selection. No reinforcement-learning or
PPO stage is used for the reported policy.

\section{LLM prompts}
\label{app:llm_prompts}

The LLM acts only as the frontier-selection policy. For any molecule selected
for expansion, the deterministic one-step engine generates the same successors
irrespective of the policy driving the search. Training and inference use the
same prompt structure, and the prompted and fine-tuned policies receive the
same system message and user-turn format. The default observation contains the
candidate SMILES and retrosynthetic graph depth; EC annotations and other
engine-derived scores are not exposed to the model.

\paragraph{System prompt.}
\begin{quote}
\small\ttfamily
You are the search policy of a retrobiosynthesis planner.

The planner works backwards from a target molecule toward metabolites available
in the host organism. At each step, a deterministic reaction engine expands one
frontier molecule using enzymatic reaction rules and adds the resulting
precursors to the search graph.

Your only task is to choose which molecule from the current frontier should be
expanded next. You do not propose reactions, select reaction rules, or evaluate
their chemical validity; these operations are handled by the deterministic
reaction engine.

The host organism is E. coli. Its sink contains roughly 750 metabolites,
including central carbon intermediates such as pyruvate, succinate, and
acetyl-CoA, amino acids, fatty acids, nucleobases, and precursors of secondary
metabolism such as chorismate.

A molecule is solved when it belongs to the sink, or when every precursor of at
least one reaction producing it is solved. There is no need to expand a
molecule further once it belongs to the sink.

Each frontier candidate is listed as:

[i] SMILES | depth=d

`depth` is the retrosynthetic graph depth of the candidate from the target.

The total number of expansions is limited and shared across the whole search
graph. Choose the frontier molecule on which the next expansion should be
spent.

Return only:

\{"choice": N\}
\end{quote}

\paragraph{User-turn template.}
At each decision, the user turn is assembled from the current search graph:
\begin{quote}
\small\ttfamily
Target molecule: <SMILES>

Search graph so far: <n> molecules, <n> reactions, <n> already reachable from
the chassis.

Frontier candidates:

[i] <SMILES> | depth=<d>

...

(<shown> of <total> frontier molecules shown)

Which candidate should be expanded next?
\end{quote}

The \texttt{depth} field is the retrosynthetic graph depth of the candidate
from the target and should not be confused with the global expansion budget.
It is the only candidate-level quantity computed by the search engine and
explicitly provided to the default LLM policy. The parenthetical truncation
line is omitted when the complete frontier is displayed.

\paragraph{Expected assistant output.}
The completion is strict JSON with a single field:
\begin{quote}
\small\ttfamily
\{"choice": <index>\}
\end{quote}
The response format is enforced through a JSON schema, so neither LLM arm can
emit prose. An unparsable answer or an index outside the displayed candidate
range is recorded as a fallback, after which the first displayed candidate is
selected.

\paragraph{Compact example.}
\begin{quote}
\small\ttfamily
Target molecule: C=C(Cl)C(=O)[O-]

Search graph so far: 5 molecules, 4 reactions, 0 already reachable from the
chassis.

Frontier candidates:

[0] C=C(Cl)C(=O)OC | depth=1

[1] CC(O)(Cl)C(=O)[O-] | depth=1

[2] CC(Cl)C(=O)[O-] | depth=1

[3] C=C(Cl)C(=O)O | depth=1

Which candidate should be expanded next?

Assistant: \{"choice": 3\}
\end{quote}

No reaction rule, EC annotation, native-substrate similarity, sink-closeness
score, tool output, or interaction history is included in the default prompt.
Each decision is a fresh exchange; the graph-summary line is the only global
state summary provided to the model.

\section{Matched-observation control}
\label{app:matched_observation}

The LLM selects only among the portfolio-defined observation
$V_t\subseteq F_t$, with $|V_t|\leq20$, whereas the classical policies
ordinarily act on the complete frontier. To test whether this asymmetry affects
end-to-end solve rate, we rerun the four classical baselines on Golden at
$N=200$ while restricting every decision to the same 20-candidate view
available to the agent. Reaction generation, graph updates, targets, expansion
budget, and policy definitions are otherwise unchanged.

\begin{table}[h]
\centering
\caption{Matched-observation control on Golden at $N=200$. Classical policies
are evaluated on the complete frontier $F_t$ and on the portfolio-defined
20-candidate view $V_t$. Exp.$\rightarrow$sol. is measured over successful
restricted-view runs.}
\label{tab:matched_observation}
\small
\begin{tabular}{lrrrr}
\toprule
Policy & Full $F_t$ & Matched $V_t$ & $\Delta$ & Exp.$\rightarrow$sol. \\
\midrule
BFS & 75\% & 75\% & 0 & 36.5 \\
DFS & 45\% & 50\% & +5 & 9.3 \\
Greedy sink similarity & 45\% & 65\% & +20 & 15.1 \\
MCTS & 80\% & 80\% & 0 & 11.9 \\
\bottomrule
\end{tabular}
\end{table}

The two strongest classical baselines are insensitive to the restriction:
MCTS and BFS retain the same solve rates on $V_t$. By contrast, both weaker
policies improve, with DFS gaining 5 percentage points and greedy sink
similarity 20. The bounded observation therefore acts as more than a smaller
frontier: the portfolio can filter out candidates favored by a poorly aligned
selection criterion while leaving already effective policies unaffected.

The large gain for greedy sink similarity is consistent with sink proximity
being a weak guide to productive retrobiosynthetic search. Restricting this
policy to the portfolio removes many of the alternatives that it would
otherwise prioritize from the complete frontier. The matched-observation
control therefore provides no evidence that access to $F_t$ disadvantages the
classical baselines relative to the LLM; if anything, the curated view can
benefit weaker selection policies. The control remains specific to Golden and
$N=200$.

\end{document}